\documentclass[10pt,letterpaper]{article}

\usepackage{cogsci}

\cogscifinalcopy %%% Uncomment this line for the final submission

\usepackage[
  style=apa,
  natbib=true,
  annotation=false,
]{biblatex}
\usepackage{float} %%% Roger Levy added this and changed figure/table placement to [H] for conformity to Word template, though floating tables and figures to top is still generally recommended!

\usepackage{amsmath}
\usepackage{booktabs}
\usepackage{graphicx}
\usepackage{subcaption}
\graphicspath{{./image/}}

\usepackage{enumitem}

\usepackage{authblk}
\usepackage[T1]{fontenc} % 确保特殊字符正常显示
\DeclareMathOperator{\Acc}{Acc}
\DeclareMathOperator{\SA}{SA}
\DeclareMathOperator{\Sens}{Sens}
\DeclareMathOperator{\Gap}{Gap}
\DeclareMathOperator{\Impact}{Impact}
\DeclareMathOperator{\Uni}{Uni}
\DeclareMathOperator{\sgn}{sgn}
\newcommand{\tfidf}{\operatorname{tf\mbox{-}idf}}
\newcommand{\pbase}{\ensuremath{p_{\mathit{base}}}}
\title{A Systematic Analysis of the Impact of Persona Steering on LLM Capabilities}

\author{
{\large\bfseries
Jiaqi Chen$^{1,*}$ \quad Ming Wang$^{1,2,*}$ \quad Tingna Xie$^{1}$ \quad Shi Feng$^{1,\dagger}$ \quad Yongkang Liu$^{3,\dagger}$
}\\
{\normalsize\normalfont
$^{1}$School of Computer Science and Engineering, Northeastern University, Shenyang 110819, China\\
$^{2}$School of Computing and Information Systems, Singapore Management University, Singapore 178902, Singapore\\
$^{3}$School of Computer and Communication Engineering, Northeastern University, Qinhuangdao 066004, China\\
$^{*}$Equal contribution \qquad
$^{\dagger}$Corresponding author: \texttt{fengshi@cse.neu.edu.cn}, \texttt{liuyongkang@qhd.neu.edu.cn}
}
}

\begin{document}

\maketitle

% \begin{abstract}
% Include no author information in the initial submission, to facilitate blind review. AI tools cannot
% be listed as authors, and authors retain full responsibility for the accuracy, integrity, and
% originality of all content in their manuscripts. This includes verifying factual claims, ensuring
% proper attribution of ideas, and confirming that the work meets standards for academic integrity and
% does not contain plagiarized content. See the Acknowledgments section of the template for AI use
% declaration and acknowledgment. The abstract should be one paragraph, no more than 150~words,
% indented 1/8~inch on both sides, in 9~point font with single spacing. The heading
% ``\textbf{Abstract}'' should be 10~point, bold, centered, with one line of space below it. This
% one-paragraph abstract section is required only for standard proceedings papers. Following the
% abstract should be a blank line, followed by the header ``\textbf{Keywords:}'' and a list of
% descriptive keywords separated by semicolons, all in 9~point font, as shown below.

% \textbf{Keywords:}
% add your choice of indexing terms or keywords;
% kindly use a semicolon;
% between each term
% \end{abstract}

\begin{abstract}

  Imbuing Large Language Models (LLMs) with specific personas is prevalent for tailoring interaction styles, yet the impact on underlying cognitive capabilities remains unexplored. We employ the Neuron-based Personality Trait Induction (NPTI) framework to induce Big Five personality traits in LLMs and evaluate performance across six cognitive benchmarks. Our findings reveal that persona induction produces stable, reproducible shifts in cognitive task performance beyond surface-level stylistic changes. These effects exhibit strong task dependence: certain personalities yield consistent gains on instruction-following, while others impair complex reasoning. Effect magnitude varies systematically by trait dimension, with Openness and Extraversion exerting the most robust influence. Furthermore, LLM effects show 73.68\% directional consistency with human personality-cognition relationships. Capitalizing on these regularities, we propose Dynamic Persona Routing (DPR), a lightweight query-adaptive strategy that outperforms the best static persona without additional training. Code and data are available at: \url{https://github.com/cjia7/DPR}.


\end{abstract}

\begin{figure*}[t]
  \centering
  \includegraphics[width=\textwidth]{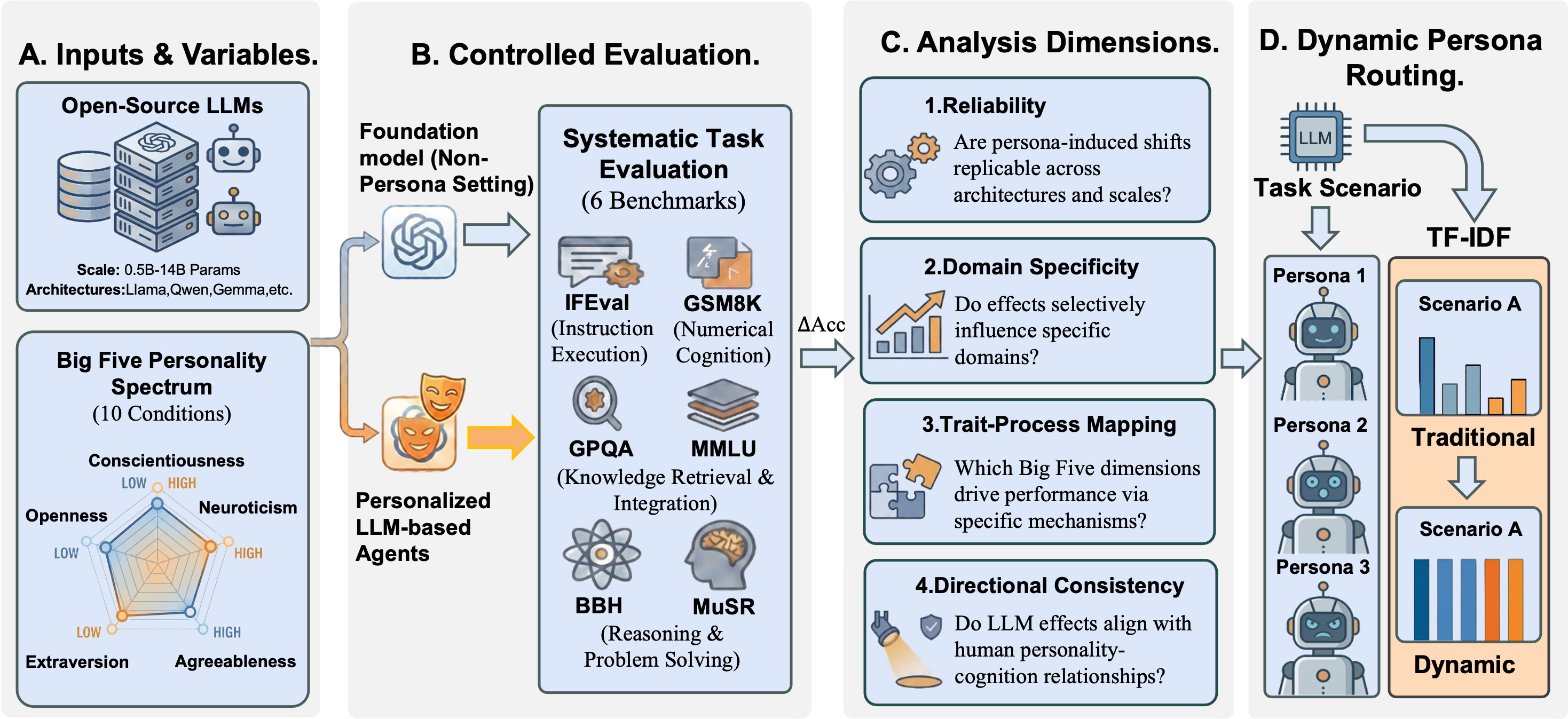}
  \caption{The systematic analysis pipeline for quantifying persona steering effects.}
  \label{fig:first}
\end{figure*}

\section{Introduction}
Personalized large language models (LLMs) have become foundational in applications demanding human-like engagement \citep{DBLP:conf/emnlp/TsengHHCHMC24,DBLP:journals/tmlr/ZhangRKSYZDB0K025}. Persona configurations tailor a model's tone and interaction style \citep{DBLP:journals/tmlr/Chen00YZSXLYZCL24}, aligning behaviors with user expectations. Research in computational personality has demonstrated that LLMs exhibit measurable behavioral patterns consistent with the Big Five framework \citep{PMID:2283588,digman1990personality,DBLP:journals/corr/abs-2306-16388}. However, whether such configurations systematically affect cognitive capabilities remains open. This study addresses: Does persona induction merely alter surface-level presentation \citep{DBLP:journals/corr/abs-2402-01765,DBLP:conf/emnlp/DeshpandeMRKN23}, or does it produce measurable shifts in cognitive task performance?

We propose a systematic framework grounded in cognitive science theory, as illustrated in Figure~\ref{fig:first}. Our pipeline comprises three stages: (1) personality trait induction via the Neuron-based Personality Trait Induction (NPTI) framework \citep{DBLP:conf/iclr/DengTYYZW25}, which modulates trait-specific neurons to induce Big Five personality configurations; (2) systematic evaluation across multiple model architectures and scales on six cognitive benchmarks \citep{DBLP:journals/jmlr/ChowdheryNDBMRBCSGSSTMRBTSPRDHPBAI23}; and (3) quantitative analysis of persona-task interactions using metrics grounded in cognitive science theory. Drawing on Cybernetic Big Five Theory (CB5T) \citep{DEYOUNG201533}, which conceptualizes traits as cybernetic control governing goal-directed behavior, we address:
\begin{itemize}[topsep=0pt, parsep=0pt, partopsep=0pt, itemsep=0pt]
  \item \textbf{RQ1 (Reliability)}: Are persona-induced performance shifts replicable across different model architectures and parameter scales, representing consistent cognitive modulation rather than stochastic noise?
  \item \textbf{RQ2 (Domain Specificity)}: Do persona effects exhibit selective influence on specific cognitive domains (reasoning, instruction-following, knowledge retrieval), consistent with the CB5T framework that predicts trait-process specificity?
  \item \textbf{RQ3 (Trait-Process Mapping)}: Which Big Five dimensions most strongly influence cognitive performance, and do they map onto distinct computational mechanisms as predicted by human personality-cognition research \citep{DEYOUNG201533,anglim2022personality}?
  \item \textbf{RQ4 (Directional Consistency)}: Do LLM persona effects align directionally with established human personality-cognition relationships, suggesting shared functional principles despite architectural differences?
\end{itemize}

Our investigation yields a critical insight: persona steering does not provide universal capability lift; rather, its efficacy is strictly conditioned on the task scenario, echoing Attentional Control Theory (ACT) \citep{eysenck2007anxiety} regarding trait-dependent cognitive resource allocation. Capitalizing on this persona-task interaction, we propose Dynamic Persona Routing (DPR), a retrieval-based strategy that adaptively applies optimal persona configurations \citep{DBLP:conf/acl/SalemiMBZ24}. DPR outperforms the best static persona baseline without additional training, establishing persona steering as a low-cost calibration mechanism.

Our contributions are:
\begin{itemize}[topsep=0pt, parsep=0pt, partopsep=0pt, itemsep=0pt]
  \item A rigorous analysis pipeline for persona steering grounded in cognitive science theory, enabling systematic quantification of how personality traits influence model capabilities across architectures.
  \item Empirical evidence for trait-process specificity: Openness and Extraversion exhibit the strongest effects, with 73.68\% directional consistency with human personality-cognition research.
  \item A training-free dynamic persona routing method demonstrating persona control as a ``low-cost calibration'' tool.
\end{itemize}

\section{Related Work}

\paragraph{Personality and Cognition in Psychology}
The relationship between personality and cognition is well-established. The Cybernetic Big Five Theory (CB5T) \citep{DEYOUNG201533} links each trait to distinct control systems: Openness to cognitive exploration, Conscientiousness to goal persistence, Extraversion to approach motivation, Agreeableness to social cooperation, and Neuroticism to threat sensitivity. Meta-analyses confirm robust associations: Openness correlates with fluid intelligence and creative problem-solving \citep{anglim2022personality}, Conscientiousness predicts academic performance \citep{fleming2016cognitive}, and Extraversion benefits social tasks. Attentional Control Theory (ACT) \citep{eysenck2007anxiety} explains how anxiety, the core of Neuroticism, impairs cognition by consuming working memory. These frameworks provide empirically validated predictions for how traits modulate cognitive processes, supporting evaluation of LLM persona effects. Furthermore, these theories emphasize the dynamic interplay between personality traits and situational demands, highlighting the importance of context in shaping cognitive outcomes.

\paragraph{Persona Induction in LLMs}
In LLM research, persona construction includes prompting-based \citep{DBLP:conf/emnlp/TsengHHCHMC24,DBLP:journals/tmlr/Chen00YZSXLYZCL24} and fine-tuning \citep{DBLP:journals/tmlr/ZhangRKSYZDB0K025} methods, mainly affecting style rather than cognition. Psychometric tools show LLMs can simulate Big Five structures \citep{digman1990personality,PMID:2283588,DBLP:journals/corr/abs-2306-16388,DBLP:journals/corr/abs-2409-15256}, but most work is descriptive, not causal. Persona assignments shift behavioral biases \citep{DBLP:conf/emnlp/DeshpandeMRKN23,DBLP:conf/acl/PerezRLNCHPOKKJ23}, but effects on reasoning remain unclear. The NPTI framework \citep{DBLP:conf/iclr/DengTYYZW25} enables neuron-level trait induction, allowing systematic quantification of personality effects while controlling for prompt confounds. Notably, this approach bridges the gap between descriptive and causal analyses, offering a robust methodology for disentangling the cognitive and stylistic dimensions of persona effects.

\paragraph{Persona Effects on Model Behavior}
Recent work shows persona conditioning interacts with instruction-tuning: personas can increase compliance and social appropriateness but may trade off with reasoning or factual consistency \citep{DBLP:journals/corr/abs-2402-01765,DBLP:conf/emnlp/DeshpandeMRKN23}. Different persona settings bias both content and style. This motivates moving beyond prompt-level personas to controlled interventions for causal analysis. NPTI \citep{DBLP:conf/emnlp/LesterAC21} provides a neuron-level mechanism for trait induction, enabling rigorous tests of whether induced personality modulates cognitive subsystems. Our work systematically maps persona effects across domains, architectures, and scales, grounded in cognitive science theory. Additionally, the interplay between persona traits and task complexity underscores the need for adaptive strategies, such as dynamic persona routing, to optimize performance across diverse scenarios.

\section{Methodology}

\subsection{Models and Cognitive Task Batteries}
To systematically assess the impact of personality on LLM capabilities, we construct a model set $\mathcal{M}$ and a task set $\mathcal{D}$ designed to control for architectural and scaling factors. We select open-source instruction-tuned models along two comparison axes: to evaluate generalizability across model families (RQ1), we employ four representative models in the 7B--9B parameter range, including LLaMA-3-8B-Instruct, Mistral-7B-v0.3, Gemma-2-9B-Instruct, and Qwen2.5-7B-Instruct; to examine scaling effects (also RQ1), we analyze the Qwen2.5 family across five scales (0.5B, 1.5B, 3B, 7B, and 14B). Performance is evaluated on six benchmarks spanning four cognitive domains (RQ2): IFEval for instruction comprehension and execution; MMLU-Pro and GPQA for knowledge retrieval and expert-level understanding; BBH and MuSR for multi-step reasoning and problem-solving; and GSM8K for numerical reasoning.

\subsection{Personality Trait Induction}
To induce personality traits without confounds from prompt engineering, we employ the Neuron-based Personality Trait Induction (NPTI) framework, which operates at the representation level by modulating specific neurons within the Feed-Forward Networks (FFN). The process involves two phases. In the \textit{identification phase}, using the PersonalityBench dataset, we compute the activation probability difference ($\delta$) for each neuron across contrasting high- and low-trait samples. Neurons with $\delta > \tau$ are identified as trait-specific, forming positive ($\mathcal{N}^+$) and negative ($\mathcal{N}^-$) sets for each Big Five dimension. In the \textit{steering phase}, we apply deterministic modulation to these neurons during inference without updating model weights. For a target persona $p$, the activation $h_i$ of neuron $i$ is modified as $h'_i = h_i + \alpha \cdot h_{ref}$ if $i \in \mathcal{N}^+$, and suppressed if $i \in \mathcal{N}^-$, where $\alpha$ is the steering strength. The baseline condition ($\pbase$) corresponds to standard inference with $\alpha=0$. Prior validation has demonstrated that this intervention reliably induces psychometrically valid personality traits, providing a rigorous causal basis for quantifying personality effects on cognitive performance.

\subsection{Experimental Procedure}

To ensure reproducibility and minimize generation noise, all models were evaluated using deterministic decoding (temperature = 0.0). We adhered to standard evaluation protocols, utilizing official prompts and scoring scripts, including few-shot configurations for BBH, GPQA, and MMLU-Pro. For GSM8K, we applied relaxed regex-based answer extraction to mitigate false negatives from formatting variations; for IFEval, we employed the strict instruction-following evaluator. Critically, to isolate the causal impact of personality, we employed a \textit{within-item paired design}: for every evaluation instance, both baseline and persona-steered conditions processed identical inputs under identical generation parameters. Thus, any performance deviation is exclusively attributable to the induced persona configuration.

\subsection{Analysis Framework}
We establish a hierarchical analysis framework to quantify persona effects. Let $\Acc(m,p,d)$ denote the accuracy of model $m$ under persona condition $p$ on dataset $d$. We include a no-persona baseline ($\pbase$) and ten polarity conditions based on the Big Five dimensions: $\mathcal{P}=\{A_H, A_L, C_H, C_L, E_H, E_L, N_H, N_L, O_H, O_L\}$, where subscript $H$ denotes high-trait and $L$ denotes low-trait (reversed) conditions. The persona effect ($\Delta \Acc$) is defined as the deviation from baseline:
\begin{equation}
  \Delta\Acc(m,p,d) = \Acc(m,p,d) - \Acc(m,\pbase,d)
\end{equation}

Positive values indicate performance gains; negative values indicate degradation. By analyzing this differential metric rather than raw accuracy, we isolate performance shifts attributable to persona intervention. $\Delta \Acc$ serves as the fundamental measure for addressing our research questions (RQ1--RQ4).

\subsubsection{Consistency Across Architectures (RQ1)}
Generalizability across model families is evaluated on a cross-architecture subset $\mathcal{M}_{a}$, which comprises models of comparable scale (7B--9B parameters). We quantify cross-architecture consistency using two complementary metrics. First, we compute the mean effect $\overline{\Delta\Acc}_{a}$ by macro-averaging $\Delta\Acc$ across all models in $\mathcal{M}_{a}$, capturing the aggregate effect size and direction (positive for gains, negative for degradation). Second, to assess whether the effect direction remains stable across architectures, we define direction consistency ($\SA$) as:
\begin{equation}
  \SA(p,d) = \frac{1}{|\mathcal{M}_{a}|}\sum_{m\in\mathcal{M}_{a}}
  \mathbb{I}\left(\sgn(\Delta\Acc) =\sgn(\overline{\Delta\Acc}_{a})\right)
\end{equation}

where $\mathbb{I}(\cdot)$ denotes the indicator function and $\sgn(\cdot)$ is the sign function. Values of $\SA$ close to 1.0 indicate that the effect is highly reproducible and largely architecture-independent.
% ($s \in {0.5\text{B}, \dots, 14\text{B}}$)//这边换了
\subsubsection{Scaling Trends (RQ1)}
We analyze how parameter scale affects persona sensitivity within the Qwen2.5 family ($s \in {\mathrm{Qwen2.5-0.5\text{B}-it}, \dots, \mathrm{Qwen2.5-14\text{B}-it}}$). To enable meaningful cross-task comparisons, we define the relative persona effect as the accuracy change normalized by baseline performance:
\begin{equation}
  \Delta \Acc_{\mathit{rel}}(s, p, d) = \frac{\Delta \Acc(s, p, d)}{\Acc(s, \pbase, d)}
\end{equation}

This preserves the direction of the effect (positive for gains, negative for degradation). For aggregate analysis, we compute persona sensitivity ($\Sens$) as the mean of absolute relative effects across all persona conditions:
\begin{equation}
  \Sens(s,d) = \frac{1}{|\mathcal{P}|} \sum_{p \in \mathcal{P}} |\Delta \Acc_{\mathit{rel}}(s, p, d)|
\end{equation}

Scaling effects are then assessed using Spearman's rank correlation ($\rho$). In particular, we examine two distinct trends: the direction trend ($\rho_{\mathit{dir}}$), computed between log-parameters and $\Delta \Acc$ to test whether larger models exhibit greater performance gains; and the sensitivity trend ($\rho_{\mathit{mag}}$), computed between log-parameters and $\Sens$ to determine whether scaling amplifies susceptibility to persona interventions.

\subsubsection{Domain Specificity (RQ2)}
We examine persona effects across cognitive domains by aggregating benchmarks into four semantic categories and computing within-type mean effects ($\overline{\Delta\Acc}(p,g)$), enabling identification of stable trait-task interaction patterns.

\subsubsection{Trait-Process Mapping (RQ3)}
We isolate the influence of specific traits by analyzing the polarity gap ($\Gap(m,t,d)$) for each dimension $t$. This metric is defined as the difference between accuracy under high ($t_H$) and low ($t_L$) settings: $\Gap(m,t,d) = \Acc(m,t_H,d) - \Acc(m,t_L,d)$. Dominance is then characterized via two metrics. The first is impact ($\Impact(t)$), measured by the mean absolute gap across all models and datasets. The second is uniformity ($\Uni(t)$), which quantifies the directional consistency relative to the global mean gap ($\overline{\Gap}(t)$):
\begin{equation}
  \Uni(t) = \frac{1}{|\mathcal{M}||\mathcal{D}|}\sum_{m\in\mathcal{M}}\sum_{d\in\mathcal{D}}
  \mathbb{I}\left(\sgn(\Gap) =\sgn(\overline{\Gap}(t))\right)
\end{equation}

High $\Uni(t)$ indicates that a specific polarity is universally advantageous regardless of the model or task.

\subsubsection{Human-LLM Directional Consistency (RQ4)}
We compare LLM persona effects against predictions from psychological literature \citep{anglim2022personality,DEYOUNG201533,eysenck2007anxiety}: Openness should enhance cognitive flexibility (high $>$ low); Conscientiousness should benefit goal-directed performance (high $>$ low); Neuroticism should impair performance (low $>$ high). We compute the directional consistency rate as the proportion of trait-benchmark combinations where observed LLM effects match predicted human patterns.

\section{Experiments}

Our evaluation reveals that persona induction produces structured and reproducible shifts in cognitive task performance rather than random noise. While baseline accuracies for all models across the six benchmarks are detailed in Table~\ref{tab:base_six_datasets_pct_once}, our analysis focuses on the relative deviation ($\Delta \Acc$) to isolate the causal impact of personality. These effects are profoundly task-dependent: personas consistently enhance instruction-following but frequently impair complex reasoning and mathematical performance. The magnitude and direction of these shifts are modulated by both parameter scale and architectural differences.
% 需要在导言区：\usepackage{booktabs}

% 导言区需要：
% \usepackage{booktabs}
% \usepackage{graphicx} % 为了 \resizebox

\begin{table}[t]
  \centering
  \caption{Base accuracy on six benchmarks (unit: \%).}
  \label{tab:base_six_datasets_pct_once}
  \resizebox{\linewidth}{!}{%
    \begin{tabular}{lcccccc}
      \toprule
      Model              & GPQA  & BBH   & MuSR  & MMLU  & IFEval & GSM8K \\
      \midrule
      \multicolumn{7}{l}{\textit{Cross-Architecture}} \\
      LLaMA-3-8B-it      & 30.13 & 63.40 & 54.10 & 36.55 & 39.93  & 72.18 \\
      Gemma-2-9B-it      & 29.02 & 70.96 & 57.67 & 50.38 & 47.87  & 58.83 \\
      Mistral-7B-it-v0.3 & 30.36 & 45.92 & 45.37 & 33.36 & 34.20  & 45.49 \\
      Qwen2.5-7B-it      & 27.68 & 66.73 & 50.40 & 55.09 & 44.55  & 81.58 \\
      \midrule
      \multicolumn{7}{l}{\textit{Scaling Analysis}} \\
      Qwen2.5-14B-it     & 31.47 & 78.16 & 62.96 & 63.05 & 41.40  & 89.58 \\
      Qwen2.5-3B-it      & 22.77 & 50.41 & 46.83 & 41.19 & 37.52  & 66.19 \\
      Qwen2.5-1.5B-it    & 27.46 & 35.57 & 45.37 & 29.11 & 21.26  & 38.44 \\
      Qwen2.5-0.5B-it    & 30.13 & 11.70 & 37.04 & 14.10 & 19.77  & 14.10 \\
      \bottomrule
    \end{tabular}%
  }
\end{table}

%\subsubsection{RQ1: Consistency across architectures} 
% \begin{figure}[htbp]
%   \centering
%   \includegraphics[width=\columnwidth]{image/heatmap_large_font.pdf}
%   \caption{Heatmaps of $\Delta\Acc_{\mathit{arch}}(p,d)$ and $\SA$}
%   \label{fig:RQ1}
% \end{figure}

\subsection{RQ1: Consistency Across Architectures}

\begin{figure}[t]
  \centering
  \begin{subfigure}[b]{\columnwidth}
    \includegraphics[width=\textwidth]{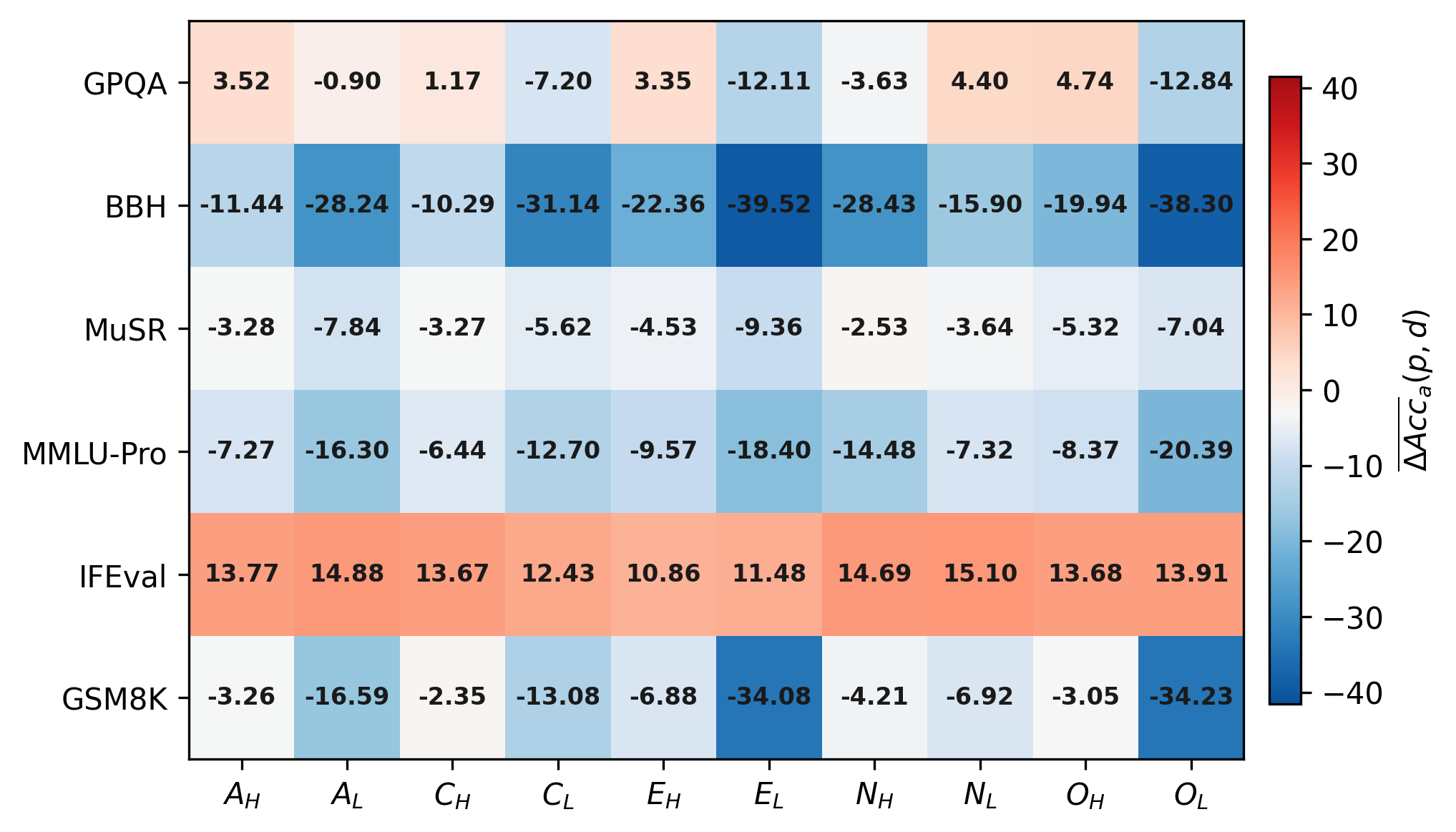}
    \caption{Signed $\overline{\Delta\Acc}_{a}$ (\%)}
    \label{fig:RQ1a}
  \end{subfigure}
  \vspace{0.3em}
  \begin{subfigure}[b]{\columnwidth}
    \includegraphics[width=\textwidth]{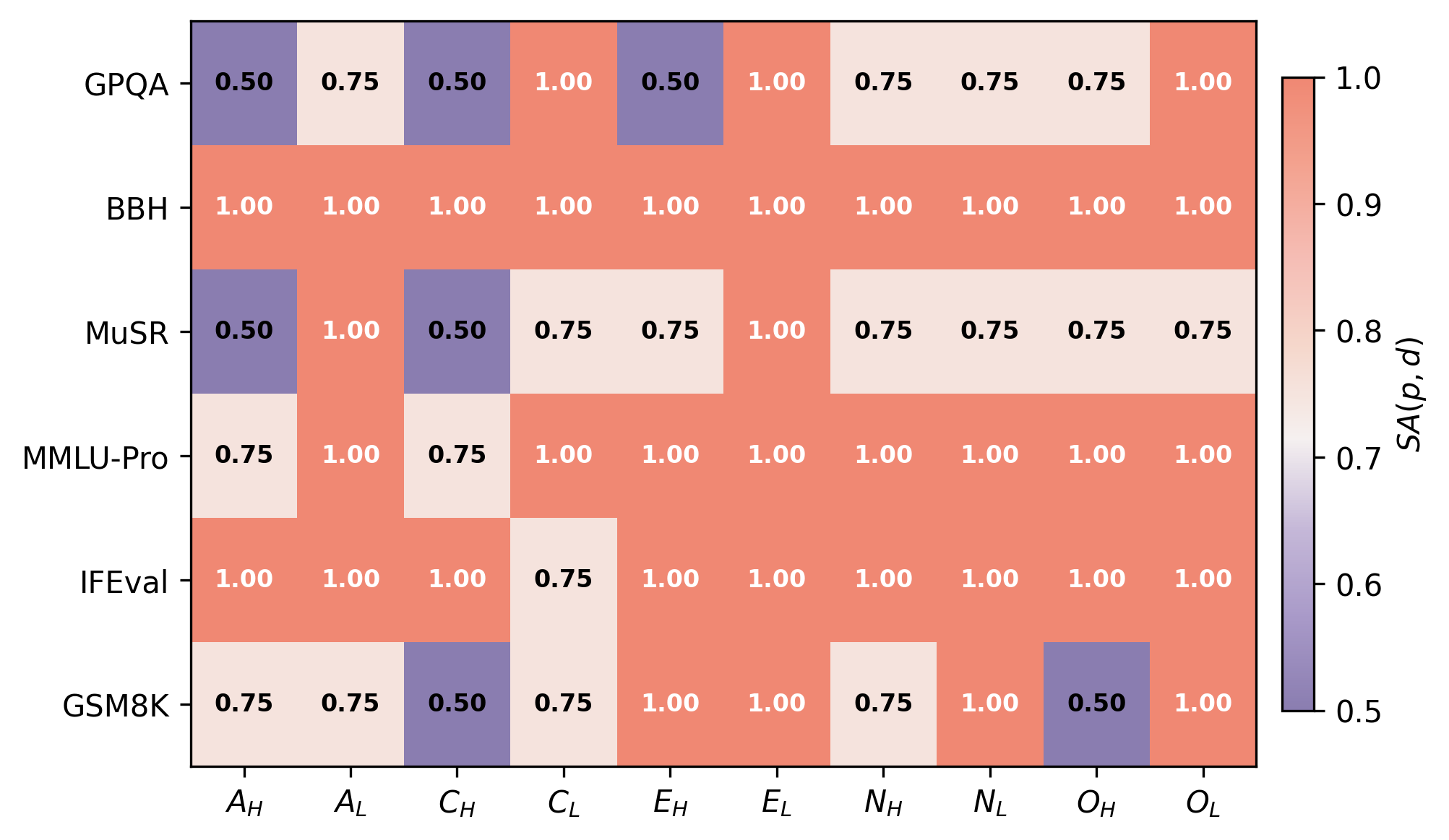}
    \caption{Direction consistency ($\SA$)}
    \label{fig:RQ1b}
  \end{subfigure}
  \caption{Persona-task interaction heatmap showing $\Delta \Acc$ and direction consistency ($\SA$).}
  \label{fig:RQ1}
\end{figure}

To determine whether persona effects reflect model-specific artifacts or deeper computational regularities, we evaluate the cross-architecture subset $\mathcal{M}_a$ (7B--9B) using mean effect ($\overline{\Delta \Acc}_{a}$) and direction consistency ($\SA$). As shown in Figure~\ref{fig:RQ1}, persona induction produces highly consistent behavioral shifts across distinct cognitive domains. The heatmap displays signed $\overline{\Delta \Acc}_{a}$ for all ten persona conditions ($A_H, A_L, \ldots, O_H, O_L$), with positive values (warm colors) indicating performance gains and negative values (cool colors) indicating degradation relative to baseline.

The near-perfect consistency ($\SA \approx 0.98$) in IFEval (all personas yield gains of $+10.9\%$ to $+15.1\%$) and BBH ($\SA = 1.00$; all personas degrade performance, with $E_L$ causing $-39.5\%$) suggests that persona induction acts as a global modulator of cognitive state. Low-trait conditions ($O_L, E_L$) consistently impair reasoning across all architectures, indicating that personality representations reconfigure shared computational mechanisms for logical operations.

In contrast, the lower consistency in GPQA ($\SA \approx 0.75$) and MuSR ($\SA \approx 0.75$) reveals a boundary: while process-oriented capabilities (reasoning, instruction-following) are governed by architecture-agnostic mechanisms, knowledge-dependent expert understanding is more tightly coupled to model-specific representational space.

\subsection{RQ2: Domain Specificity}

\begin{figure}[t]
  \centering
  \begin{subfigure}[b]{\columnwidth}
    \includegraphics[width=\textwidth]{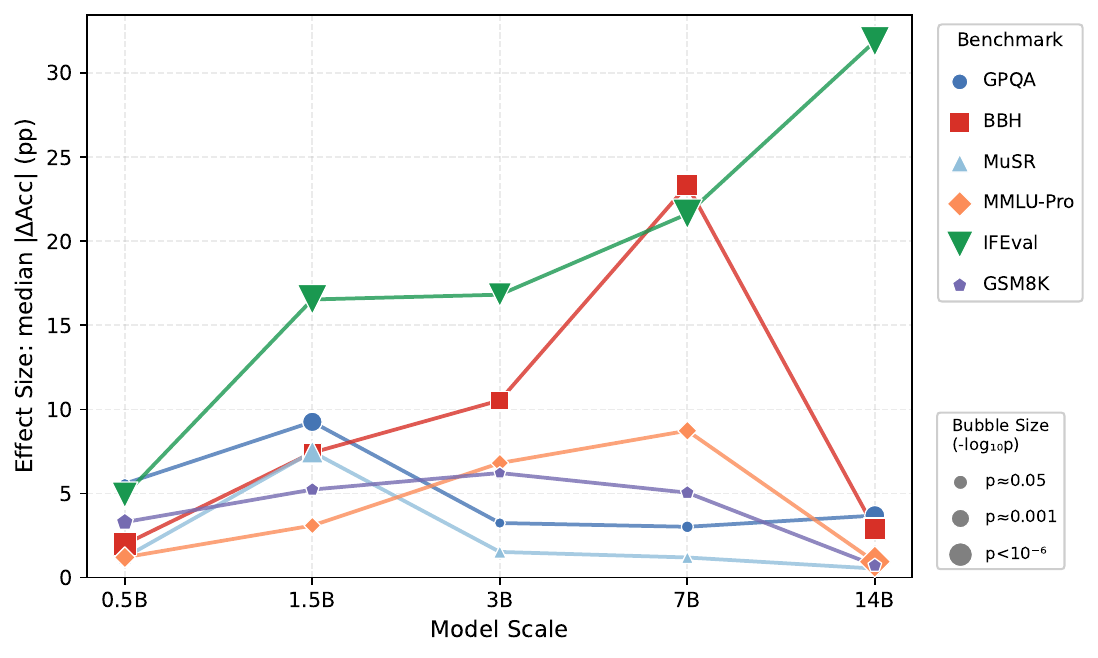}
    \caption{Effect size (median $|\Delta\Acc|$, pp) across model scales. Bubble size encodes statistical evidence: larger bubbles indicate stronger significance ($-\log_{10}p$).}
    \label{fig:RQ2a}
  \end{subfigure}
  \vspace{0.3em}
  \begin{subfigure}[b]{\columnwidth}
    \includegraphics[width=\textwidth]{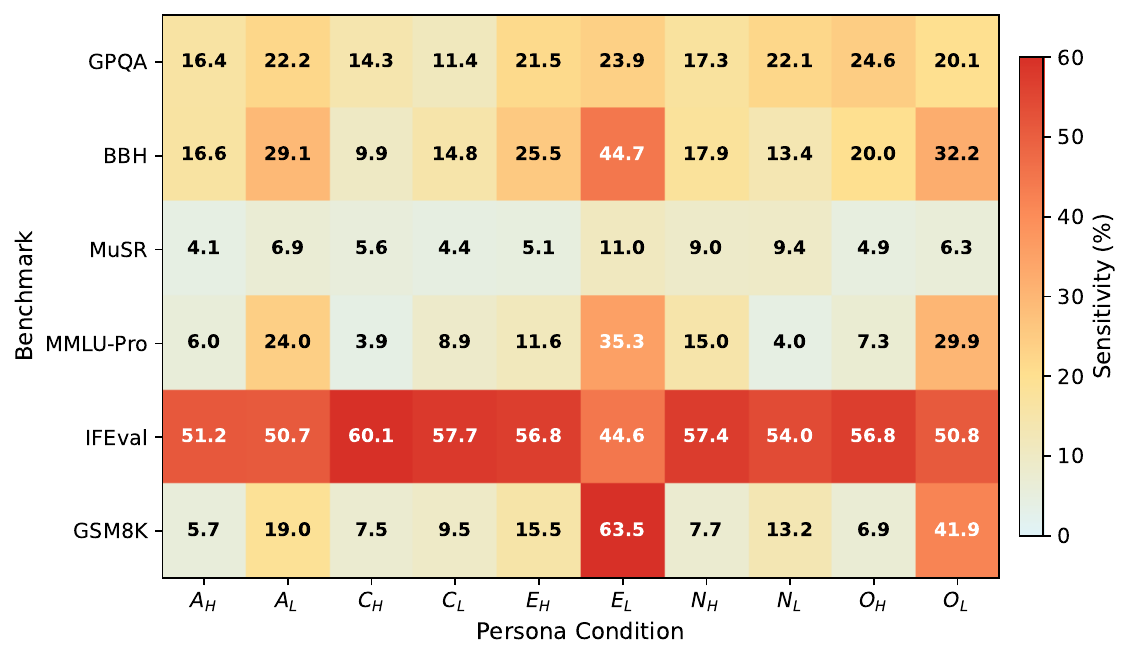}
    \caption{Persona-specific sensitivity by task}
    \label{fig:RQ2b}
  \end{subfigure}
  \caption{Domain-specific persona effects in Qwen2.5 family (0.5B--14B). Panel~(a) integrates effect magnitude (line height) and statistical evidence (bubble size) to reveal both the strength and reliability of persona effects across scales.}
  \label{fig:RQ2}
\end{figure}

We investigate how persona effects vary across cognitive domains, with an additional lens on parameter scaling (0.5B to 14B) within the Qwen2.5 family. As shown in Figure~\ref{fig:RQ2}, persona effects exhibit clear domain selectivity consistent with CB5T's prediction of trait-process specificity. Panel~(a) presents a dual-metric visualization: the $y$-axis shows the robust effect size, median absolute accuracy change across persona conditions (in percentage points), while bubble size encodes statistical evidence strength ($-\log_{10}p$, from paired $t$-tests comparing persona and baseline accuracies). Larger bubbles indicate stronger evidence that persona effects are non-random. Panel~(b) provides a fine-grained breakdown of sensitivity by individual persona condition.
For instruction-following (IFEVal), both effect size and significance remain consistently high across scales, indicating sustained responsiveness to persona-driven behavioral modulation. Conversely, reasoning-intensive benchmarks (BBH, GSM8K) show non-monotonic trajectories: effect sizes peak at 7B but attenuate sharply at 14B, where small bubbles indicate that residual effects approach noise levels. This pattern suggests that larger models develop more robust reasoning schemas increasingly invariant to persona interventions.

These patterns reveal a developmental dissociation: stylistic flexibility in social-communicative tasks emerges with linguistic complexity, while reasoning capabilities consolidate into schemas increasingly invariant to persona interventions, aligning with human findings that personality traits differentially impact distinct cognitive systems \citep{DEYOUNG201533}.

\begin{table}[b]
  \centering
  \caption{Impact and Uniformity Across Big Five Dimensions}
  \label{tab:dimension_analysis}
  \resizebox{\columnwidth}{!}{%
  \begin{tabular}{lcccc}
    \toprule
    \textbf{Trait} & \textbf{Impact} & \textbf{Avg. Gap} & \textbf{Uniformity} & \textbf{Rank} \\
    \midrule
    Openness (O)          & 11.96\% & +11.17\% & 90.5\% & 1 \\
    Extraversion (E)      & 11.70\% & +10.42\% & 90.5\% & 1 \\
    Agreeableness (A)     & 8.06\%  & +6.50\%  & 73.8\% & 4 \\
    Conscientiousness (C) & 6.31\%  & +5.82\%  & 88.1\% & 3 \\
    Neuroticism (N)       & 4.00\%  & $-$1.86\%  & 57.1\% & 5 \\
    \bottomrule
  \end{tabular}%
  }
\end{table}
\subsection{RQ3: Trait-Process Mapping}
Table~\ref{tab:dimension_analysis} compares dimensions by effect magnitude (Impact) and directional stability (Uniformity) using the high--low polarity gap. \textbf{Openness} and \textbf{Extraversion} emerge as dominant traits: they induce the largest effects (Impact: 11.96\% and 11.70\%) and show the most stable direction across models and tasks (Uniformity: 90.5\%). This aligns with CB5T predictions linking Openness to cognitive exploration and Extraversion to approach motivation \citep{DEYOUNG201533}. \textbf{Conscientiousness} exhibits smaller but highly stable effects (Impact: 6.31\%; Uniformity: 88.1\%), consistent with its theoretical role in goal maintenance. \textbf{Agreeableness} shows moderate magnitude with lower stability (Impact: 8.06\%; Uniformity: 73.8\%). \textbf{Neuroticism} has the weakest and least stable influence (Impact: 4.00\%; Uniformity: 57.1\%) and is the only dimension favoring the low setting (mean Gap: -1.86\%), consistent with ACT predictions that anxiety impairs cognitive efficiency \citep{eysenck2007anxiety}.

\subsection{RQ4: Human-LLM Directional Consistency}
We systematically compared LLM persona effects against predictions derived from psychological literature. Based on CB5T and ACT frameworks, we formulated directional hypotheses: Openness should enhance cognitive flexibility (high $>$ low); Conscientiousness should benefit goal-directed performance (high $>$ low); Neuroticism should impair performance (low $>$ high); Extraversion should facilitate approach-oriented tasks (high $>$ low); and Agreeableness effects should be task-dependent.
Across trait-benchmark combinations, LLM effects showed \textbf{73.68\% directional consistency} with human patterns (14/19 comparisons). Openness exhibited the highest consistency (87.5\%), with high-O outperforming low-O on 7/8 benchmarks, mirroring the Openness-intelligence association in humans ($r \approx .35$) \citep{anglim2022personality}. Conscientiousness also showed strong consistency (87.5\%), aligning with its role in goal persistence and sustained attention. Neuroticism showed lower consistency (57.1\%), though notably, low-N consistently outperformed high-N on anxiety-sensitive tasks requiring sustained attention, consistent with ACT predictions \citep{eysenck2007anxiety}.

This substantial alignment suggests that personality constructs in LLMs capture functional regularities parallel to human trait-cognition relationships. The convergence is particularly striking given that NPTI-induced traits emerge from activation manipulation rather than the developmental and biological processes underlying human personality.

\section{Leveraging Persona-Task Regularities}

The preceding analyses reveal that persona effects are strongly task-dependent rather than uniformly beneficial or detrimental. To exploit these structural regularities without additional training, we propose Dynamic Persona Routing (DPR), a lightweight retrieval-based strategy that adapts persona configurations to specific input queries.

\paragraph{Reference Construction} For each benchmark dataset $\mathcal{D}$, we partition the data into a reference set $\mathcal{R}$ and a test set $\mathcal{T}$ with a 9:1 ratio using LLaMA-3-8B-Instruct. The reference set serves as a ``routing memory,'' storing the historical performance of all persona conditions on known instances.

\paragraph{Similarity-Based Retrieval} For a given test instance $x \in \mathcal{T}$, we identify the most semantically similar historical instance (anchor) $x^*$ from $\mathcal{R}$. We employ TF-IDF vectorization to map text inputs to vector space representations, with the anchor selected via cosine similarity maximization:
\begin{equation}
x^* = \underset{r \in \mathcal{R}}{\arg\max} \frac{\tfidf(x) \cdot \tfidf(r)}{\|\tfidf(x)\| \|\tfidf(r)\|}
\end{equation}
This ensures that routing decisions are grounded in contextually relevant prior experience.

\paragraph{Routing Strategy} We assume that queries with high semantic similarity share similar sensitivities to persona steering. Based on the retrieved anchor $x^*$, we construct an Effective Persona Set $\mathcal{P}_{eff}(x^*) = \{ p \in \mathcal{P} \mid y^{(x^*, p)} = 1 \}$, the subset of persona conditions under which the model successfully solved the anchor instance. This set serves as the dynamic recommendation for the current query.

\paragraph{Evaluation} We define a ``hit'' if the recommended set contains at least one persona capable of solving $x$, comparing against $\Acc_{\mathit{best}}$ (best single fixed persona applied globally).

% preamble 里需要：
% \usepackage{booktabs}

\begin{table}[htbp]
  \centering
  \caption{Dynamic Persona Routing vs. Best Static Baseline}
  \label{tab:application-results}
  \resizebox{\columnwidth}{!}{%
    \begin{tabular}{lccccc}
      \toprule
      Dataset  & Total & Sampled & Correct & Accuracy (\%) & Best Baseline (\%) \\
      \midrule
      GPQA     & 448   & 44      & 23      & 52.27         & 41.96              \\
      BBH      & 6511  & 651     & 404     & 62.06         & 63.40              \\
      MuSR     & 756   & 75      & 59      & 78.67         & 54.10              \\
      MMLU-Pro & 12032 & 1203    & 519     & 43.14         & 36.67              \\
      IFEVal   & 541   & 54      & 34      & 62.96         & 62.48              \\
      GSM8K    & 1319  & 131     & 88      & 67.17         & 72.18              \\

      \bottomrule
    \end{tabular}%
  }
\end{table}

\paragraph{Results} As shown in Table~\ref{tab:application-results}, dynamic persona selection outperforms the optimal static baseline on most benchmarks. The retrieval-based approach yields the most significant gains on MuSR (+24.57\%) and GPQA (+10.31\%), demonstrating that semantically similar questions do share persona sensitivities. Moderate improvements appear on MMLU-Pro (+6.47\%) and IFEVal (+0.48\%). However, on reasoning-intensive tasks (GSM8K, BBH), the dynamic strategy slightly underperforms the best static configuration, suggesting that retrieved personas may occasionally introduce interference in pure logical reasoning where consistency is paramount.

\paragraph{Analysis} The differential effectiveness across domains aligns with our earlier findings: tasks requiring flexible knowledge retrieval and multi-step reasoning (MuSR, GPQA) benefit most from adaptive persona selection, while tasks with more rigid solution structures (GSM8K mathematical reasoning) favor stable configurations. This pattern reinforces the domain-specificity principle established in RQ2.

This proof-of-concept demonstrates that even without training a dedicated routing model, the simple heuristic of ``effective personas from similar questions'' yields substantial gains on knowledge-intensive and multi-step reasoning tasks, establishing persona control as a lightweight, training-free mechanism for behavioral calibration.

\section{Conclusion}
This study systematically evaluates how Big Five traits influence LLM cognitive performance. Four key regularities emerge: (1) \textbf{Reliability}: persona effects are consistent across architectures (7B--9B), indicating shared computational mechanisms; (2) \textbf{Domain specificity}: personas enhance instruction-following but often impair reasoning, consistent with ACT predictions; (3) \textbf{Trait-process mapping}: Openness and Extraversion exert the strongest influence, as predicted by CB5T; (4) \textbf{Human-LLM alignment}: 73.68\% directional consistency with human trait-cognition relationships.
The substantial convergence suggests that NPTI-induced traits capture functional regularities analogous to biological cognition, implying architecture-independent computational structures underlying personality-cognition relationships. Openness and Extraversion's dominant influence aligns with CB5T predictions linking these dimensions to cognitive exploration and approach motivation, while Neuroticism uniquely favoring low settings mirrors ACT predictions regarding anxiety-induced attentional interference.
The Dynamic Persona Routing strategy demonstrates that persona induction functions as a behavioral hyperparameter for training-free performance calibration, with substantial gains on knowledge-intensive tasks.

\section{Acknowledgments}

The work is supported by the National Natural Science Foundation of China (Nos. 62272092, 62502081) and the Northeastern University Student Innovation and Entrepreneurship Program (No. 260192). Furthermore, we would also like to thank \href{https://kinamind.org}{the KinaMind society} for their inspiring environment and unwavering support.

\printbibliography[title={References}]

\end{document}